\documentclass[a4paper]{article}

\usepackage{INTERSPEECH2021}
\usepackage{amsmath,amssymb,graphicx,amsfonts,cite,tabularx,multirow,xcolor,url,tipa,breakurl,bm}
\usepackage[htt]{hyphenat}

\usepackage{xcolor}
\usepackage{amsmath}
\usepackage{makecell}

\newcolumntype{D}{>{\centering\arraybackslash}m{12ex}}
\newcolumntype{E}{>{\centering\arraybackslash}m{9.3ex}}
\newcolumntype{F}{>{\centering\arraybackslash}m{19.7ex}}
\newcolumntype{G}{>{\centering\arraybackslash}m{10ex}}

\title{Contextualized Streaming End-to-End Speech Recognition with\\Trie-Based Deep Biasing and Shallow Fusion}
\name{Duc Le, Mahaveer Jain, Gil Keren, Suyoun Kim, Yangyang Shi, Jay Mahadeokar, Julian Chan,\\Yuan Shangguan, Christian Fuegen, Ozlem Kalinli, Yatharth Saraf, Michael L. Seltzer}
\address{Facebook AI}
\email{duchoangle@fb.com}

\begin{document}

\maketitle
\begin{abstract}

How to leverage dynamic contextual information in end-to-end speech recognition has remained an active research area. Previous solutions to this problem were either designed for specialized use cases that did not generalize well to open-domain scenarios, did not scale to large biasing lists, or underperformed on rare long-tail words. We address these limitations by proposing a novel solution that combines shallow fusion, trie-based deep biasing, and neural network language model contextualization. These techniques result in significant 19.5\% relative Word Error Rate improvement over existing contextual biasing approaches and 5.4\%--9.3\% improvement compared to a strong hybrid baseline on both open-domain and constrained contextualization tasks, where the targets consist of mostly rare long-tail words. Our final system remains lightweight and modular, allowing for quick modification without model re-training.

\end{abstract}
\noindent\textbf{Index Terms}: end-to-end speech recognition, deep biasing, shallow fusion, contextualization, sequence transducer

\section{Introduction}
\label{sec:intro}

End-to-end automatic speech recognition (ASR) models have become increasingly popular in recent years, thanks to their simplicity and competitive performance on generic transcription tasks \cite{Prabhavalkar17,He2019RNNT,Gulati2020conformer,Zhang2021benchmark}. Meanwhile, traditional hybrid systems based on Hidden Markov Model and Deep Neural Network (HMM-DNN) still remain attractive in practice due to their modularity and flexibility. Specifically, the stand-alone language model (LM) allows for easy integration of external knowledge sources, such as unpaired text data and contextual biasing information.

Much of the past research in end-to-end ASR has focused on methods to inject external knowledge into the system. For static knowledge (i.e., knowledge that does not change from one utterance to the next), various LM fusion techniques have been proposed \cite{gulcehre2015using,Kannan2018,sriram2018cold,toshniwal2018comparison,shan2019component,Kim2021lmfusion}, in addition to methods that remove the internal LM's contribution prior to fusion \cite{McDermott19,Variani2020hat,Meng2021ILME}. For dynamic knowledge (i.e., utterance-specific contextual information), possible solutions include shallow fusion with class-based Weighted Finite State Transducer (WFST) \cite{Zhao2019,He2019RNNT,Le2021deepshallow}, attention-based deep context \cite{Pundak2018DC,Chen2019DC,Jain2020DC}, and trie-based deep biasing \cite{Jain2020DC,Le2021deepshallow}. The shallow fusion-based solutions \cite{Zhao2019,He2019RNNT,Le2021deepshallow} rely on strong context prefixes (e.g., \emph{call}, \emph{play}, \emph{message}) to increase precision and avoid overbiasing. While effective for the targeted tasks, this method does not generalize well to open-domain use cases where such prefixes are not available. On the other hand, deep biasing solutions that inject contextual information directly into the network \cite{Pundak2018DC,Chen2019DC,Jain2020DC,Le2021deepshallow} are not reliant on context prefixes or known patterns, but do not work well when the biasing list gets larger and tend to underperform on rare words. How to develop an end-to-end ASR system that can handle large biasing lists of rare words in an open-domain scenario remains a challenge.

We tackle this problem by proposing a novel solution that combines WFST shallow fusion, trie-based deep biasing, and neural network LM (NNLM) contextualization. Unlike previous work which relied on specialized WFSTs \cite{Zhao2019,He2019RNNT,Le2021deepshallow}, we use a generic WFST in this work that does not require strong context prefixes nor domain-specific sentence patterns. Unlike previous work which only contextualized the end-to-end ASR model \cite{Pundak2018DC,Chen2019DC,Jain2020DC,Le2021deepshallow}, we propose to fuse the NNLM with trie-based deep biasing to give the former implicit access to biasing information and leverage the vast amount of unpaired text data. On LibriSpeech, our proposed techniques produce on average \textbf{34.5\%} relative Word Error Rate (WER) improvement compared to the baseline and \textbf{19.5\%} compared to existing contextual biasing approaches. Finally, our experiments on large-scale in-house data validate the findings on LibriSpeech; the resulting end-to-end system improves significantly over the baseline as well as outperforms a strong hybrid setup by \textbf{5.4\%--9.3\%} on both open-domain and constrained contextualization tasks, where the targets consist of mostly rare long-tail words. Our final system remains lightweight and modular, allowing for quick modification without re-training the main ASR model.

\section{Data}
\label{sec:data}

\subsection{LibriSpeech}
\label{ssec:librispeech}

We conduct the majority of our experiments on the widely used LibriSpeech \cite{panayotov2015librispeech} dataset which consists of 960 hours of labeled audio for training and an additional 810M-word text corpus for LM building. In order to study contextualization performance, we construct an artificial biasing list for each test utterance as follows\footnote{\url{https://github.com/facebookresearch/fbai-speech/tree/master/is21_deep_bias}}. Firstly, we fill the biasing list with all \textbf{\emph{rare}} words in the reference, defined as words that fall outside the 5,000 most common words in the audio training set (which account for 90\% of all word occurrences). Secondly, we add $N=\{100, 500, 1000, 2000\}$ distractors randomly sampled from the 209.2K rare words in the training vocabulary. In evaluation, the model has access to both the audio as well as the associated biasing list. We will track three different evaluation metrics, (1) \textbf{\emph{WER}}: overall WER measured on all words, (2) \textbf{\emph{U-WER}}: unbiased WER measured on words NOT IN the biasing list, and (3) \textbf{\emph{B-WER}}: biased WER measured on words IN the biasing list. Insertion errors are counted toward B-WER if the inserted word is in the biasing list and U-WER otherwise. The biasing lists cover 5.8K out of 52.6K words in \emph{test-clean} and 5.3K out of 52.3K words in \emph{test-other} (i.e., \textbf{10.6\%} total word coverage). The goal of contextualization is to improve B-WER without degrading U-WER significantly.

\subsection{Large-Scale In-House Data}
\label{ssec:in_house_data}

The LibriSpeech corpus is significantly smaller than our typical in-house datasets. Moreover, each artificial biasing list has at least some overlap with the reference text, which is not guaranteed in practice. We therefore conduct experiments on our in-house data to understand the system's performance on large-scale training sets and realistic evaluation data. The first part of our training corpus includes 1.7M hours of English video data publicly shared by Facebook users; all videos are completely de-identified before transcription. The second part contains 50K hours of manually transcribed de-identified English data with no user-identifiable information (UII) in the voice assistant domain, similar to those described in \cite{Le2021deepshallow}. All utterances are morphed when researchers manually access them to further de-identify the speaker. Note that the data are NOT morphed during training and evaluation.

We consider two evaluation sets, (1) \textbf{\emph{Video}}: 14.0K manually transcribed de-identified English videos publicly shared by Facebook users and (2) \textbf{\emph{Assistant}}: 20.8K manually transcribed de-identified utterances collected from voice activity of volunteer participants interacting with the Facebook voice assistant. The participants consist of households that have consented to having their voice activity reviewed and analyzed. The biasing lists in Video contain on average 11 words (stdev $\delta$=18) extracted from each video's title and description, covering 30.0K out of 2.0M total reference words (i.e., \textbf{1.5\%} coverage). The biasing lists in Assistant contain on average 876 ($\delta$=491) contact names, covering 8.1K out of 60.6K total reference words excluding the wakeword (i.e., \textbf{13.4\%} coverage). Video data reflect the open-domain use case where the biasing word can appear in any context, whereas Assistant data reflect the constrained use case where the biasing word appears mostly in limited context (by itself or after \emph{call}). We will also track WER, U-WER, and B-WER for each test set.

\section{Methods}
\label{sec:methods}

\subsection{WFST Shallow Fusion with Dynamic Replacement}
\label{ssec:fst_shallow_fusion}

\begin{figure}[tb]
  \centering
  \includegraphics[width=\columnwidth]{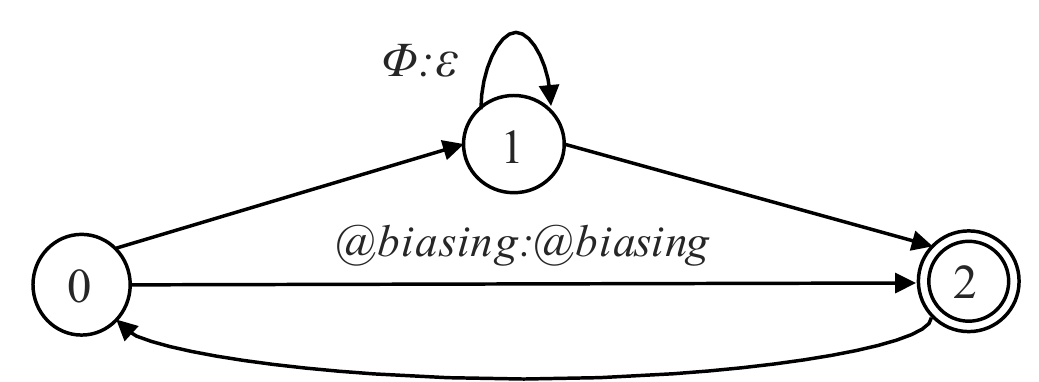}
  \caption{Generic biasing WFST. @biasing is a dynamic class and $\Phi$ is a special ilabel that can match any input symbol.}
  \label{fig:fst}
  \vspace{-1em}
\end{figure}

WFST shallow fusion applied at the decoder level is considered the standard contextualization method for end-to-end ASR \cite{Zhao2019,He2019RNNT,Le2021deepshallow}. Unlike our previous work which employed a 4-gram WFST LM trained on domain-specific sentence patterns \cite{Le2021deepshallow}, we use a simple and generically applicable WFST in this work, as shown in Figure \ref{fig:fst}. In decoding, the \texttt{@biasing} WFST is constructed dynamically from biasing data, followed by determinization, minimization, and epsilon removal. Similar to \cite{Zhao2019,Le2021deepshallow}, we apply biasing at the WordPiece level and before the pruning stage of decoding. Note that the WFST shown in Figure \ref{fig:fst} can be easily modified to incorporate context prefixes (e.g., \emph{call}, \emph{play}, \emph{message}) without losing its generalizability.

\subsection{Trie-Based Deep Biasing}
\label{ssec:deep_biasing}

\begin{figure}[tb]
  \centering
  \includegraphics[width=\columnwidth]{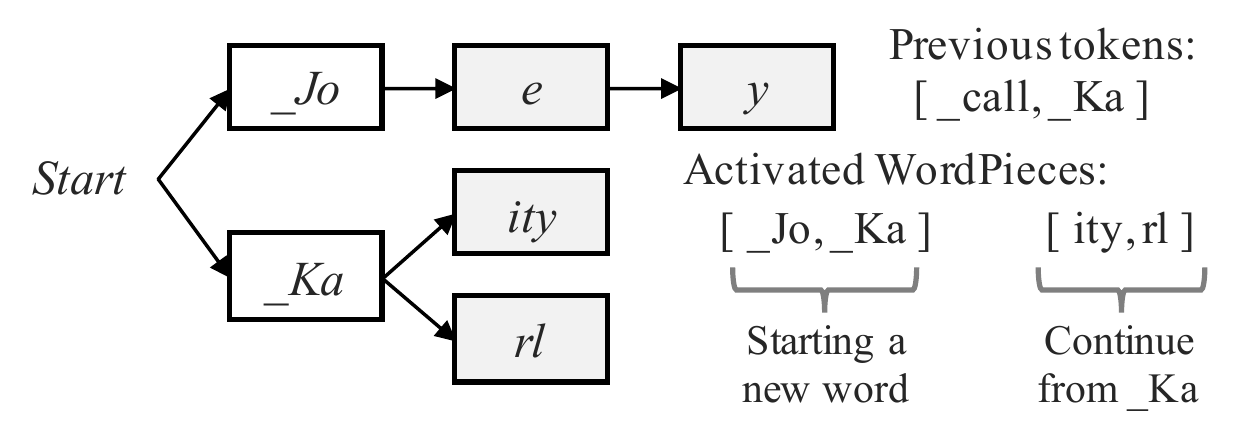}
  \vspace{-2em}
  \caption{Example trie-based biasing module given biasing words [Joe, Joey, Kaity, Karl] and the result of a sample query. Shaded cells indicate potential end-of-word symbols.}
  \label{fig:biasing_module}
\end{figure}

\begin{figure}[tb]
  \centering
  \includegraphics[width=\columnwidth]{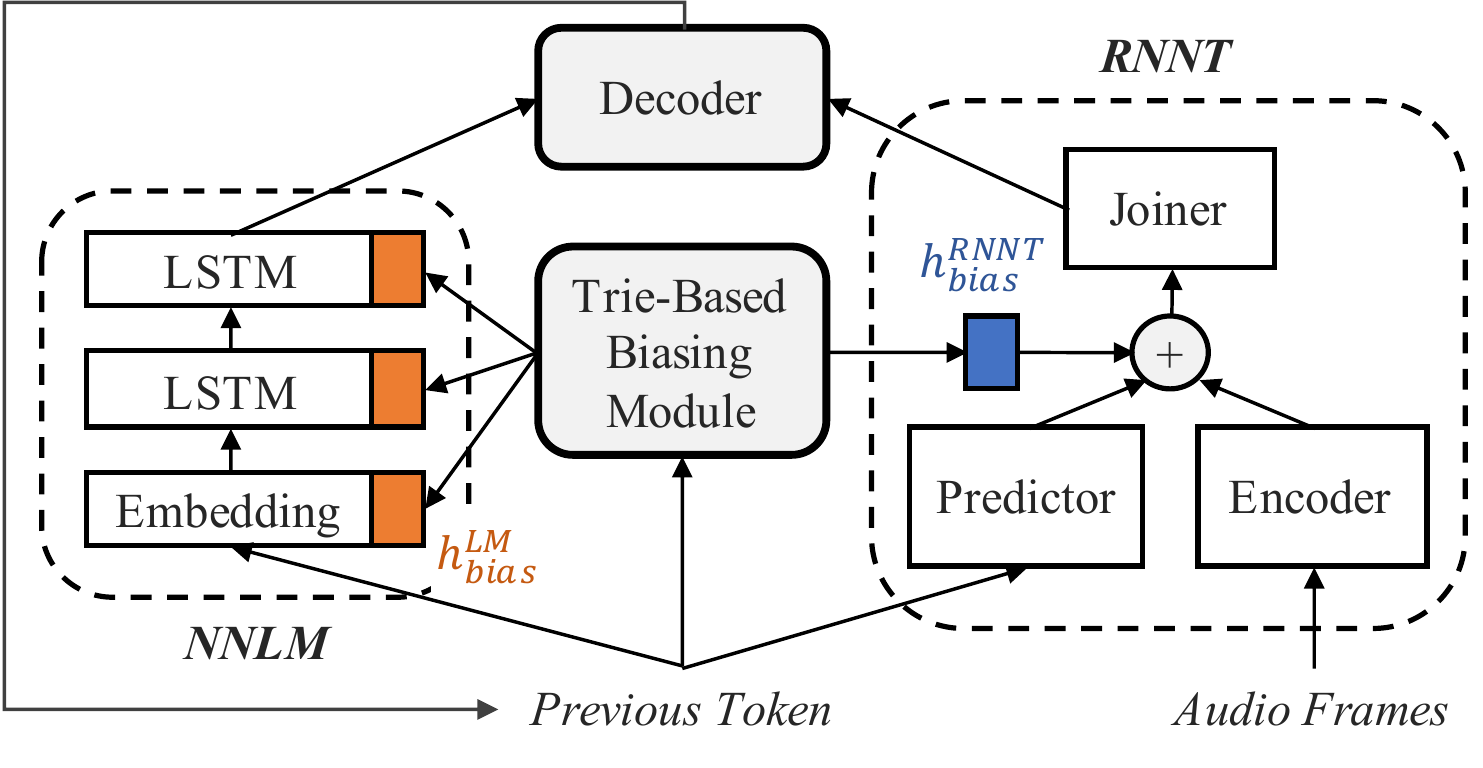}
  \caption{How the trie-based biasing module interacts with RNNT and NNLM. Shaded blocks are non-trainable.}
  \label{fig:deep_biasing}
  \vspace{-1em}
\end{figure}

The trie-based biasing module was proposed in our previous work \cite{Le2021deepshallow,Jain2020DC} to encode a list of biasing words as a non-trainable trie data structure over their WordPiece representation. In this work, we construct the trie in the same way, but slightly simplify how it is queried. Given previously emitted WordPiece tokens as query, this biasing module returns a binary vector $h_{\text{bias}} = [h_{\text{start}};h_{\text{continue}}] \in \{0,1\}^{2D}$, where $D$ is the WordPiece vocabulary size. $h_{\text{start}} \in \{0,1\}^D$ assigns a value of one to (i.e., \emph{activates}) WordPieces that can \emph{start a new word} in the biasing list, and zero to everything else. $h_{\text{continue}} \in \{0,1\}^D$ activates WordPieces that can \emph{continue an unfinished word} in the biasing list. In short, $h_{\text{bias}}$ represents all WordPieces that can continue traversing the trie given previously emitted tokens. Figure \ref{fig:biasing_module} shows an example trie and its query operation. In practice, we implement the trie as a lookup table to minimize its overhead during training and decoding; querying the trie takes $O(1)$ time and building the trie takes $O(N)$ time (negligible in practice), where $N$ is the total number of WordPieces in the biasing list.

\subsubsection{Deep Biasing RNNT (DB-RNNT)}
\label{sssec:deep_biasing_rnnt}

All of our end-to-end ASR models in this paper are variants of Sequence Transducer \cite{Graves12transduction}, also known as Recurrent Neural Network Transducer (RNNT). A core limitation of vanilla RNNT in contextualization, even with WFST shallow fusion, is that the model does not have access to any biasing information during inference. As a result, the RNNT output probabilities, especially for rare long-tail words, may be too small for decoder-side biasing to recover from. The motivation for integrating deep biasing with RNNT is to address this limitation by giving the latter implicit access to biasing data during inference. As shown in the right side of Figure \ref{fig:deep_biasing}, the output of the biasing module gets projected to the same dimension $D^{\text{emb}}$ as the RNNT's encoder and predictor embeddings, $h_{\text{bias}}^{\text{RNNT}} = f_{\text{bias}}^{\text{RNNT}}(h_{\text{bias}})$, with $f_{\text{bias}}^{\text{RNNT}}: \{0,1\}^{2D} \rightarrow \mathbb{R}^{D^{\text{emb}}}$. The projected $h_{\text{bias}}$ becomes an extra input to the joiner and the parameters of $f_{\text{bias}}^{\text{RNNT}}$ are trained jointly with the RNNT; see \cite{Le2021deepshallow} for a more detailed formulation. The joiner's job in this extended model is to integrate information from all three lower-level components, the encoder, predictor, and biasing module.

\subsubsection{Deep Biasing NNLM (DB-NNLM)}
\label{sssec:deep_biasing_nnlm}

NNLM has a similar limitation as RNNT in that without access to biasing information, the model may not assign high-enough probabilities for rare words to be recovered with only decoder-side biasing. We propose to integrate deep biasing with NNLM as well to address this limitation. As shown in the left side of Figure \ref{fig:deep_biasing}, the output of the biasing module gets projected to a smaller dimension $D^{\text{lm}}$, $h_{\text{bias}}^{\text{LM}} = f_{\text{bias}}^{\text{LM}}(h_{\text{bias}})$, with $f_{\text{bias}}^{\text{LM}}: \{0,1\}^{2D} \rightarrow \mathbb{R}^{D^{\text{lm}}}$. The projected $h_{\text{bias}}^{\text{LM}}$ gets concatenated with every hidden layer output in the NNLM and the parameters of $f_{\text{bias}}^{\text{LM}}$ are optimized together with NNLM parameters. One advantage of fusing deep biasing with NNLM compared to RNNT is that we can train the former on text-only data, which are typically much easier to obtain and more plentiful than paired audio-text data. We train the NNLM separately from RNNT in this work, which makes the system more modular. Note that it is possible to train both models jointly (e.g., with cold fusion \cite{sriram2018cold,Kim2021lmfusion}), which we do not consider here. We will describe how DB-RNNT and DB-NNLM are trained in Section \ref{sec:experiments}.

\section{Experimental Setup}
\label{sec:experiments}

\subsection{LibriSpeech Experiments}
\label{ssec:librispeech_exps}

The baseline streaming RNNT architecture used in this work has approximately 83M parameters in total. The encoder is a 20-layer streamable low-latency Emformer model \cite{Shi2021emformer} with a stride of six, 60ms lookahead, 300ms segment size, 512-dim input, 2048-dim hidden size, eight self-attention heads, and 1024-dim fully-connected (FC) projection. The predictor consists of three Long Short Term Memory (LSTM) layers with 512-dim hidden size, followed by 1024-dim FC projection. The joiner contains one Rectified Linear Unit (ReLU) and one FC layer. The target units are 5000 unigram WordPieces \cite{Kudo2018SubWord} trained with SentencePiece \cite{kudo-richardson-2018-sentencepiece}. The model is trained for 100 epochs using sub-word regularization ($l=5$, $\alpha=0.25$) \cite{Kudo2018SubWord} and AR-RNNT loss \cite{Mahadeokar2021AR-RNNT} with left buffer 0 and right buffer 15, where the alignment is provided by a chenone hybrid acoustic model (AM) \cite{Le2019Kulfi}. We utilize the auxiliary training tasks proposed in \cite{Liu2021MTL} to improve model performance and convergence, as well as speed perturbation \cite{Ko2015AudioAF} and SpecAugment LD policy \cite{park2019specaugment}.

The baseline NNLM consists of five LSTM layers with 2048-dim hidden size and 1024-dim internal projection, totaling approximately 105M parameters. The model is trained on the 810M-word text-only corpus for 40 epochs with Cross Entropy loss similar to \cite{Kim2021lmfusion}. This NNLM is shallow fused with the RNNT after discounting internal LM scores \cite{Meng2021ILME}.

For DB-RNNT and DB-NNLM training, the projection modules $f_{\text{bias}}^{\text{RNNT}}$ and $f_{\text{bias}}^{\text{LM}}$ both contain a single FC layer with output dimension 1024 and 256, respectively. We construct an artificial biasing list for each training utterance on-the-fly as described in Section \ref{ssec:librispeech}, where the number of distractors ranges from 400 to 800 (uniformly sampled). To prevent the model from overfitting to the biasing module's output, we zero out the entire $h_{\text{bias}}$ binary vector with 10\% probability and remove each target reference word from the biasing list with 40\% probability. DB-RNNT and DB-NNLM follow the same training strategy as the baseline RNNT and NNLM without deep biasing.

\begin{table*}[t]
\centering
\begin{tabular}{ | D || E | E || E | E || E | E || E | E | }
    \hline
    \multirow{2}{*}{\textbf{Model}} & \multicolumn{2}{c||}{$\mathbf{N=100}$} & \multicolumn{2}{c||}{$\mathbf{N=500}$} & \multicolumn{2}{c||}{$\mathbf{N=1000}$} & \multicolumn{2}{c|}{$\mathbf{N=2000}$} \\
    \cline{2-9}
    & \textbf{\emph{test-clean}} & \textbf{\emph{test-other}} & \textbf{\emph{test-clean}} & \textbf{\emph{test-other}} & \textbf{\emph{test-clean}} & \textbf{\emph{test-other}} & \textbf{\emph{test-clean}} & \textbf{\emph{test-other}} \\
    \hline
    \hline
    \texttt{B1}: RNNT Baseline & 3.65 (2.4/14.1) & 9.61 (7.2/30.6) & 3.65 (2.4/14.1) & 9.61 (7.2/30.6) & 3.65 (2.4/14.1) & 9.61 (7.2/30.6) & 3.65 (2.4/14.1) & 9.61 (7.2/30.6) \\
    \hline
    \texttt{S1}: \texttt{B1} + DB-RNNT & 3.11 (2.3/9.8) & 8.79 (7.1/23.4) & 3.24 (2.3/10.7) & 9.03 (7.2/25.1) & 3.30 (2.4/11.0) & 9.12 (7.2/26.1) & 3.34 (2.3/11.4) & 9.28 (7.3/27.0) \\
    \hline
    \texttt{S2}: \texttt{B1} + WFST & 3.06 (2.3/9.4) & 8.60 (7.1/22.2) & 3.10 (2.3/9.6) & 8.72 (7.1/22.5) & 3.11 (2.3/9.7) & 8.78 (7.2/22.8) & 3.09 (2.3/9.6) & 8.83 (7.2/22.9) \\
    \hline
    \textbf{\texttt{S3}: \texttt{S2} + DB-RNNT} & \textbf{2.81 (2.2/7.4)} & \textbf{8.10 (7.0/17.7)} & \textbf{2.91 (2.3/8.1)} & \textbf{8.30 (7.1/19.1)} & \textbf{3.00 (2.3/8.5)} & \textbf{8.45 (7.1/20.5)} & \textbf{3.04 (2.3/8.9)} & \textbf{8.75 (7.3/21.8)} \\
    \hline
    \hline
    \texttt{B2}: \texttt{B1} + NNLM & 2.79 (1.7/11.6) & 7.35 (5.2/26.3) & 2.79 (1.7/11.6) & 7.35 (5.2/26.3) & 2.79 (1.7/11.6) & 7.35 (5.2/26.3) & 2.79 (1.7/11.6) & 7.35 (5.2/26.3) \\
    \hline
    \texttt{S4}: \texttt{S3} + NNLM & 2.28 (1.6/7.9) & 6.50 (5.1/18.7) & 2.35 (1.6/8.2) & 6.64 (5.2/19.6) & 2.40 (1.7/8.4) & 6.72 (5.2/20.2) & 2.41 (1.7/8.6) & 6.81 (5.2/20.9) \\
    \hline
    \textbf{\texttt{S5}: \texttt{S3} + DB-NNLM} & \textbf{1.98 (1.5/5.7)} & \textbf{5.86 (4.9/14.1)} & \textbf{2.09 (1.6/6.2)} & \textbf{6.09 (5.1/15.1)} & \textbf{2.14 (1.6/6.7)} & \textbf{6.35 (5.1/17.2)} & \textbf{2.27 (1.6/7.3)} & \textbf{6.58 (5.2/18.9)} \\
    \hline
\end{tabular}
\vspace{0.5em}
\caption{LibriSpeech results with different biasing list size $N$. Reported metrics are in the following format: WER (U-WER/B-WER).}
\vspace{-1.5em}
\label{table:librispeech_results}
\end{table*}

\begin{figure}[tb]
  \centering
  \includegraphics[width=\columnwidth]{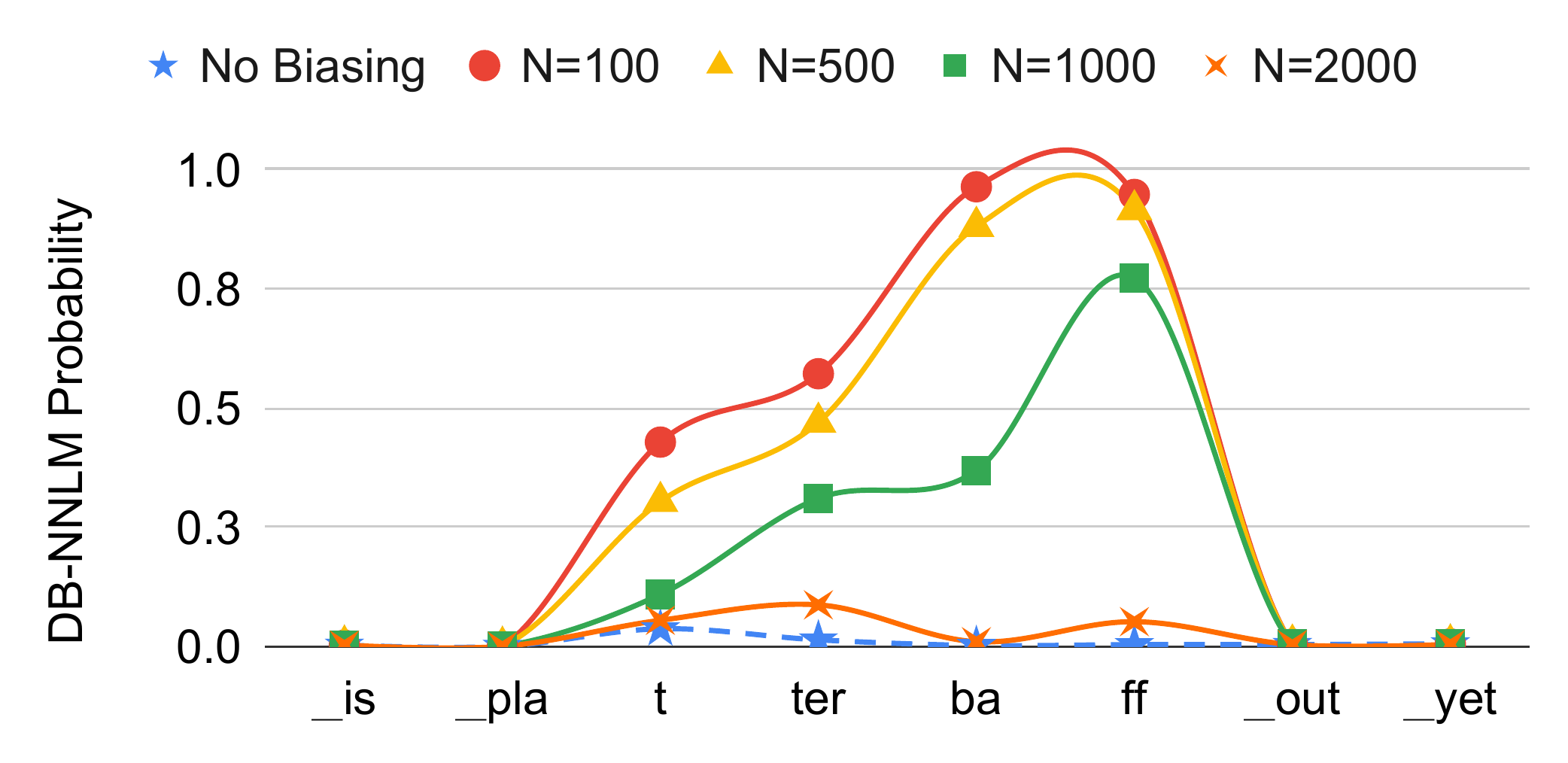}
  \caption{LibriSpeech DB-NNLM probability with different biasing list size $N$. ``platterbaff" is one of the biasing words.}
  \label{fig:lmprob}
  \vspace{-1em}
\end{figure}

\subsection{In-House Data Experiments}
\label{ssec:in_house_exps}

Our experiments on in-house data employ similar model architectures and training strategies described above with several differences. Firstly, the DB-RNNT follows a two-stage training process similar to \cite{Le2021deepshallow}, where the joiner parameters are reset before finetuning the entire model with a lower learning rate. This helps reduce experiment turnaround time given the large amount of training data. Secondly, unlike LibriSpeech where the text-only corpus is significantly larger than the paired audio-text data (approximately 86 times the number of words), our in-house NNLM training corpus is only 20\% larger than the paired audio-text portion. This may reduce the potential WER improvement from NNLM and DB-NNLM. Thirdly, we utilize G2G \cite{Le2020G2G} during training and decoding to generate graphemic pronunciation variants for biasing words, which has been shown to improve rare word recognition significantly \cite{Le2020G2G,Le2021deepshallow}.

\section{Results and Discussion}
\label{sec:results_and_discussion}

\subsection{LibriSpeech Results}
\label{ssec:librispeech_results}

As shown in Table \ref{table:librispeech_results}, all of our proposed biasing techniques incur minimal degradation on U-WER, thus we focus exclusively on B-WER in the analysis. The first section of the table shows the results without NNLM. Similar to \cite{Le2021deepshallow}, we found that WFST shallow fusion performs better than DB-RNNT by itself (\texttt{S2} vs. \texttt{S1}). DB-RNNT is also quite sensitive to the size of the biasing list, where the B-WER improvement compared to the baseline \texttt{B1} gets progressively smaller as the biasing list gets larger. WFST shallow fusion exhibits a similar trend but to a much lesser extent. This behavior is due to how the DB-RNNT is trained, where we sample between 400-800 distractors for each training iteration; as a result, the model does not generalize well to much larger biasing lists. We found that sampling more distractors during training improves B-WER on larger biasing lists, but degrades B-WER on smaller biasing lists. This suggests that the number of distractors sampled during training should be based on the expected biasing list size in testing. We will investigate ways to make this technique more robust to biasing list sizes in future work. Combining DB-RNNT and WFST leads to significant \textbf{13.4\%} relative improvement in B-WER over the best standalone technique and \textbf{38.5\%} over the RNNT baseline \texttt{B1}, averaged over all test sets and biasing lists.

The second section of Table \ref{table:librispeech_results} shows the results after incorporating NNLM shallow fusion. Similar to \cite{Kim2021lmfusion}, we observe significant WER improvement over the baseline without NNLM. However, comparing \texttt{S4} and \texttt{S3}, the impact of NNLM on B-WER is overall neutral since the model does not have access to any biasing information. By replacing NNLM with DB-NNLM (\texttt{S5}), we can improve B-WER significantly as well, with an average relative improvement of \textbf{19.5\%} over \texttt{S3} and \textbf{34.5\%} over the RNNT + NNLM baseline \texttt{B2}. DB-NNLM has a similar weakness as DB-RNNT in that it does not work as well with biasing lists that are much larger than what is seen during training. Figure \ref{fig:lmprob} clearly illustrates this phenomenon, where the NNLM assigns a relatively low probability to the rare word ``platterbaff" without biasing. With ``platterbaff" in the biasing list, however, its probability becomes much higher and reduces to the non-biased probability as the biasing list gets larger. One notable observation is that the DB-NNLM's probability for ``platterbaff" is low for the first WordPiece, but gets gradually higher within the word. We observe the same trend with DB-RNNT's internal LM, thus we do not show it in this paper. By contrast, the WFST boosting weight is constant for every WordPiece in the biasing word. This suggests that: (1) the low initial probability may explain why trie-based deep biasing by itself underperforms compared to WFST shallow fusion, and (2) there may be a better way to combine the two techniques by taking advantage of the score patterns. We will explore these directions in future work to further improve the system.

\subsection{In-House Data Results}
\label{ssec:in_house_results}

\begin{table}[t]
\centering
\begin{tabular}{ | F | G | G | }
    \hline
    \multicolumn{1}{|c|}{\textbf{Model}} & \textbf{Video} & \textbf{Assistant} \\
    \hline
    \hline
    \texttt{B0}: Hybrid Baseline (\emph{with biasing}) & 16.75 (17.3/16.6) & 10.07 (9.7/12.9) \\
    \hline
    \texttt{B1}: RNNT Baseline (\emph{without biasing}) & 16.08 (16.5/27.2) & 14.36 (9.1/48.3) \\
    \hline
    \hline
    \texttt{S1}: \texttt{B1} + DB-RNNT & 16.02 (16.5/22.1) & 12.41 (9.1/33.6) \\
    \hline
    \texttt{S2}: \texttt{B1} + WFST & 16.01 (16.6/19.8) & 9.29 (8.8/12.8) \\
    \hline
    \texttt{S3}: \texttt{S2} + DB-RNNT & 16.00 (16.6/16.8) & 9.25 (8.9/11.8) \\
    \hline
    \texttt{S4}: \texttt{S3} + NNLM & 15.59 (16.2/16.7) & 9.16 (8.8/11.7) \\
    \hline
    \textbf{\texttt{S5}: \texttt{S3} + DB-NNLM} & \textbf{15.58 (16.2/15.7)} & \textbf{9.16 (8.8/11.7)} \\
    \hline
\end{tabular}
\vspace{0.5em}
\caption{In-house results. Metrics: WER (U-WER/B-WER).}
\vspace{-2em}
\label{table:in_house_results}
\end{table}

Table \ref{table:in_house_results} summarizes the results on our large-scale in-house data. Overall, we observe a similar trend as on LibriSpeech, where our final system \texttt{S5} significantly improves B-WER by \textbf{42.3\%--75.8\%} with minimal U-WER degradation, compared to the RNNT baseline \texttt{B1}. Our system also achieves \textbf{5.4\%--9.3\%} improvement on both U-WER and B-WER over the strong hybrid baseline \texttt{B0} which utilizes chenone-based Emformer AM with similar size as the RNNT encoder, 4-gram class-based LM for first pass decoding and contextual biasing, and NNLM for second pass rescoring, while being much more lightweight in physical size (200MB vs. 4GB, both with 8-bit quantization). These results confirm that our proposed techniques generalize well to large-scale training sets and realistic evaluation data.

At the same time, there are several differences compared to LibriSpeech results. Firstly, the impact of DB-RNNT and DB-NNLM is larger on Video than on Assistant, most likely because the former has smaller biasing lists which are beneficial for trie-based deep biasing. Secondly, the incremental improvement from DB-NNLM in both sets is much smaller than that in LibriSpeech. As mentioned in Section \ref{ssec:in_house_exps}, our NNLM training data is not vastly larger than the paired audio-text data used for RNNT training, which could explain the smaller improvement. This suggests that DB-NNLM is more applicable when we have limited labeled audio relative to text-only data.

\section{Conclusion and Future Work}

In this work, we proposed a novel approach to end-to-end ASR contextual biasing that combines trie-based deep biasing, WFST shallow fusion, and NNLM contextualization. Our final system achieves significant WER improvement compared to previous approaches on both academic and large-scale in-house data, and outperforms a strong hybrid baseline which has much higher disk and memory footprint. We also publicly share our experimental setup on LibriSpeech to encourage replication and wider adoption of the proposed techniques. For future work, we plan to investigate methods to make trie-based deep biasing more robust to different biasing list sizes and close the performance gap with WFST shallow fusion. We will also extend the deep biasing technique beyond contextualization use cases to improve generic speech recognition accuracy.

\section{Acknowledgment}
\label{sec:acknowledgment}

We'd like to thank Antoine Bruguier, Rohit Prabhavalkar, Frank Seide, and Xuedong Zhang for their advice and suggestions.

\bibliographystyle{IEEEtran}
\bibliography{mybib}

\end{document}